\documentclass[runningheads]{llncs}

\usepackage{eccv}

\usepackage{eccvabbrv}
\usepackage{xcolor}
\usepackage{colortbl}
\usepackage{graphicx}
\usepackage{booktabs}
\usepackage{float} 
\usepackage[numbers]{natbib}
\usepackage[accsupp]{axessibility}  

\usepackage[pagebackref,breaklinks,colorlinks,citecolor=eccvblue]{hyperref}

\usepackage{orcidlink}

\begin{document}

\title{Hybrid Latents - \\ Geometry-Appearance-Aware Surfel Splatting} 

\titlerunning{Hybrid Latents}

\author{Neel Kelkar\inst{1,2}\orcidlink{1234-5678-9012}, 
Simon Niedermayr\inst{1}\orcidlink{0009-0008-3370-0149}, 
Klaus Engel\inst{2}\orcidlink{0009-0001-1423-898X},
Rüdiger Westermann\inst{1}\orcidlink{0000-0002-3394-0731}}

\authorrunning{N.~Kelkar et al.}

\institute{Technical University of Munich, Munich, Germany\\
\email{\{neel.kelkar, simon.niedermayr, westermann\}@tum.de} \and
Siemens Healthineers, Erlangen, Germany\\
\email{\{neel.kelkar, engel.klaus\}@siemens-healthineers.com}}

\maketitle
\begin{figure}
  \centering
    \begin{minipage}[t]{0.7\textwidth}
        \vspace{-1em}
        \includegraphics[width=\textwidth]{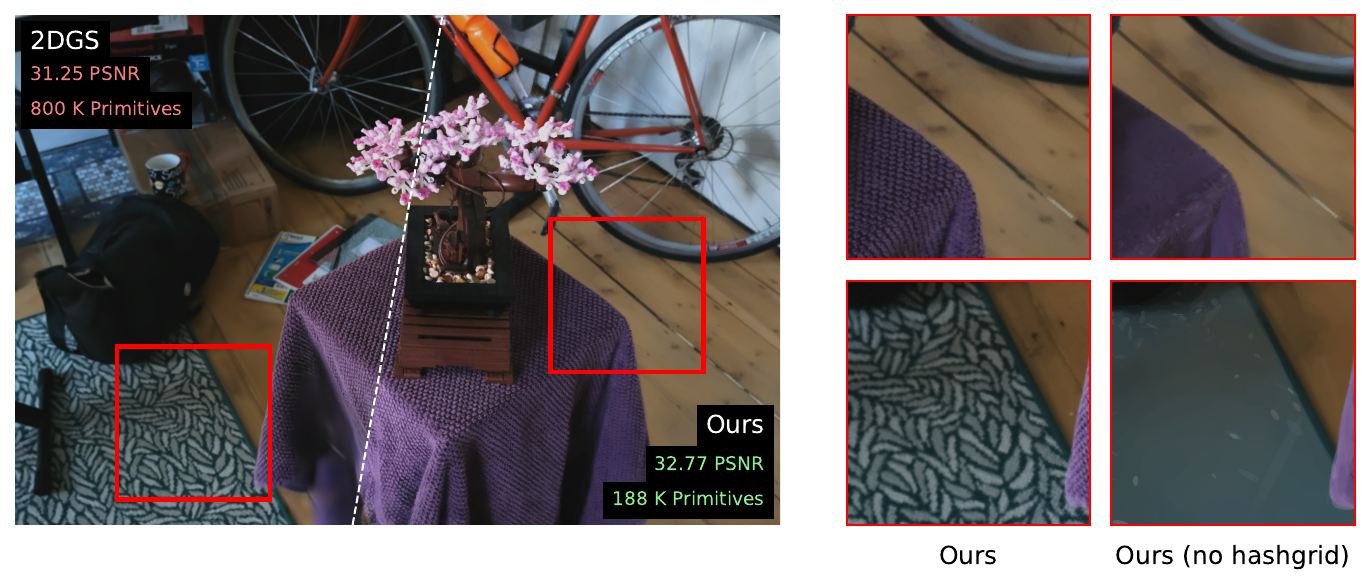}
    \end{minipage}
    \hfill
    {\color{black!20}\vrule width 0.5pt} 
    \hfill
    \begin{minipage}[t]{0.26\textwidth}
        \vspace{0pt}
        \includegraphics[width=\textwidth]{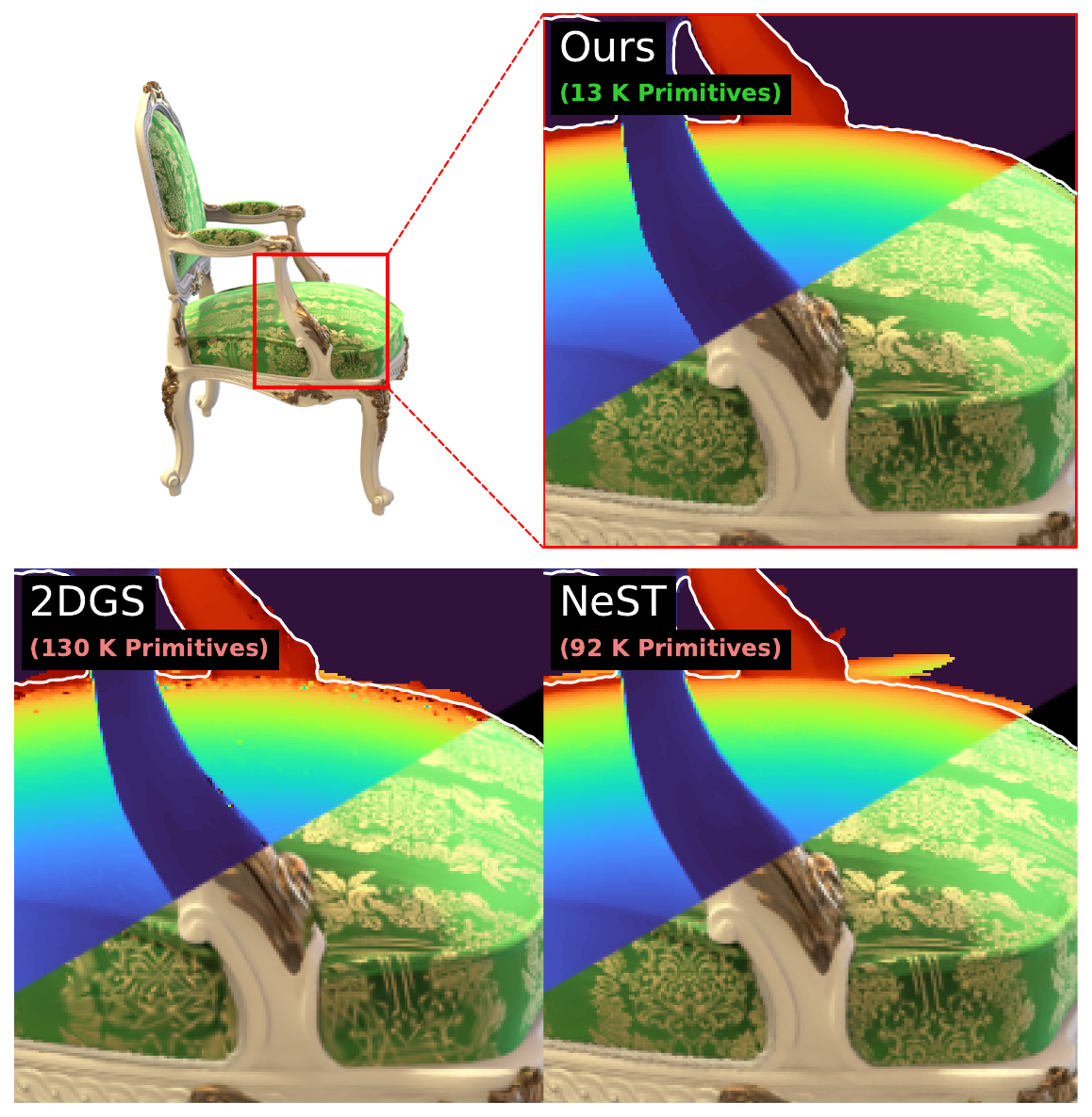}
    \end{minipage}
    \caption{\textbf{Hybrid Latents} disentangle low-frequency scene components (via per-surfel latent features) from high-frequency texture details (via a hash-grid). They achieve superior visual quality with fewer surfels and improve geometric fidelity, shown by an accurate silhouette (vs. ground truth in white) and depth reconstruction (right).}
  \label{fig:teaser}
  \vspace{-3em} 
\end{figure}

\begin{abstract}
  We introduce a hybrid Gaussian-hash-grid radiance representation for reconstructing 2D Gaussian scene models from multi-view images. Similar to NeST splatting, our approach reduces the entanglement between geometry and appearance common in NeRF-based models, but adds per-Gaussian latent features alongside hash-grid features to bias the optimizer toward a separation of low- and high-frequency scene components. This explicit frequency-based decomposition reduces the tendency of high-frequency texture to compensate for geometric errors. Encouraging Gaussians with hard opacity falloffs further strengthens the separation between geometry and appearance, improving both geometry reconstruction and rendering efficiency. Finally, probabilistic pruning combined with a sparsity-inducing BCE opacity loss allows redundant Gaussians to be turned off, yielding a minimal set of Gaussians sufficient to represent the scene. Using both synthetic and real-world datasets, we compare against the state of the art in Gaussian-based novel-view synthesis and demonstrate superior reconstruction fidelity with an order of magnitude fewer primitives.
\end{abstract}

\section{Introduction}
\label{sec:intro}

Neural Radiance Fields (NeRF) \cite{mildenhall_nerf_2021} and
3D Gaussian Splatting (3DGS) \cite{kerbl_3d_2023} have 
fundamentally redefined the benchmarks for high-fidelity 3D model reconstruction and photorealistic Novel View Synthesis (NVS). While NeRF rests upon a rigorous, continuous volumetric framework that for the first time achieved unprecedented detail in synthesizing new viewpoints from sparse 2D images, 3DGS 
shifted the paradigm towards explicit primitives, enabling real-time rasterization and compatibility with traditional graphics pipelines. 

Subsequent works have explored alternative primitive kernels to improve reconstruction quality and efficiency. 2D Gaussian Splatting (2DGS) \cite{Huang2DGS2024} utilizes oriented planar Gaussian discs as surface elements (surfels), with ray-surfel intersection to precisely model scene geometry. However, because each Gaussian encodes appearance via a single set of Spherical Harmonics (SH), 2DGS requires a prohibitive number of primitives to capture high-frequency texture variations, leading to high memory costs and redundant geometry.

To mitigate this problem, high-frequency texture variations can be modelled explicitly. 
Textures can be encoded into the primitives \cite{chao2025textured,rong2025gstex,svitov2025billboard,papantonakis2025content}, but this suffers from memory restrictions and unstable optimization. Neural Shell Texture Splatting (NeST) \cite{zhang2025neural} 
advances the 3DGS-2DGS transition by using 2D Gaussians and offloading a scene's appearance to a hash-grid-based neural feature field. Such a hybrid representation encourages a cleaner separation between geometry and texture, simultaneously reducing the number of primitives and their required capacity. 

However, because the 2D Gaussians serve merely as "coordinate samplers" for the neural field, the optimization often bloats the geometry into a dilated volumetric hull solely to use high-capacity appearance features that compensate for geometric errors (see Fig.~\ref{fig:teaser}). 
Geometry-related low-frequency scene components, which are smooth in space and stable across viewpoints, cannot be encoded efficiently in a hash-grid without overfitting. This results in reduced accuracy due to inaccurate surface reconstruction and slower rendering from an excessive number of primitives.

We propose a geometry-appearance-aware decoupling that restores the geometric role of the surfels. We introduce a \textit{hybrid latent} representation where each 2D surfel carries a base feature signal that replaces the coarse levels of the hash-grid. This forces the surfels to learn the low-frequency "base coat" of the scene geometry and lighting, while the finer layers of the hash-grid focus on high-frequency residual textures. This inductive bias stabilizes optimization, as the per-surfel component compensates for the unstable positional gradients caused by discontinuous hash collisions. This prevents geometric dilation and allows the primitives to snap tightly to the true surface, as shown in Fig. \ref{fig:teaser}. By reducing the number of required primitives and avoiding unnecessary hash-grid queries, we achieve real-time rendering speeds faster than other primitive texturing approaches.

To encourage a tight geometric representation, we replace standard Gaussian kernels with bounded Beta kernels \cite{liu2025deformable}. These kernels can adaptively morph between soft volumetric blobs and hard, planar disks with a compact support.
This allows our method to efficiently capture flat opaque surfaces while minimizing overdraw and to utilize softer surfels to model complex surfaces and volumetric components.


To optimize these primitives, we leverage a stochastic MCMC framework \cite{kheradmand20243d,liu2025deformable}, which replaces traditional heuristic-based densification~\cite{kerbl_3d_2023} with a principled relocation strategy. This approach treats primitive opacities as probabilities, allowing "dead" surfels in under-reconstructed regions to be respawned in active areas of the scene. However, while MCMC ensures robust primitive placement, it can yield a "foggy" volumetric distribution, increasing the number of redundant hash-grid queries.

To resolve this, we introduce a Binary Cross-Entropy (BCE) loss on opacity values to penalize semi-transparency. This optimization aggressively eliminates redundant primitives, achieving significantly sparser reconstructions than prior hybrid methods. Furthermore, this high degree of sparsity, combined with the compact support of the beta kernel that enables efficient axis-aligned culling \cite{wang2024adr}, minimizes expensive neural queries in empty space and maintains real-time rendering performance.

In summary, our contributions are:
\begin{itemize}
\item A hybrid latent decomposition that reduces the entanglement of geometry and appearance. This delivers improved surface accuracy, increased training and rendering speed, and reduces the required number of surfels.

\item Utilizing Beta kernels to minimize overdraw and maximize reconstruction sparsity wherever possible. This further boosts rendering speed without compromising reconstruction quality.

\item A tailored optimization framework that integrates Bayesian sampling (MCMC) with sparsity induction. This results in a robust pipeline that achieves superior sparsity, handles reconstruction ambiguities, and maintains real-time rendering performance.
\end{itemize}

\section{Related Work}


\paragraph{\textbf{Novel View Synthesis}}
Neural Radiance Fields (NeRF)~\cite{mildenhall_nerf_2021} and its successors have significantly advanced NVS by representing scenes as continuous volumetric functions parameterized by coordinate-based neural networks. 
They achieve photorealistic quality by optimizing density and directional color via differentiable volume rendering~\cite{weiss_differentiable_2022}.
However, the high computational cost of querying MLPs along the rays limits real-time performance. 
To address this, Instant-NGP~\cite{muller_instant_2022} introduced multi-resolution hash encodings for neural features, enabling training and inference in near real time by reducing the MLP size and leveraging efficient memory lookups. Crucially, these methods model appearance as a continuous spatial function, in stark contrast to primitive-based approaches where features are usually unchanging across the extent of a discrete primitive.

\textbf{\textit{Splatting Based Approaches}}
3D Gaussian Splatting (3DGS)~\cite{kerbl_3d_2023} has emerged as a powerful explicit representation. 
By explicitly representing the scene with anisotropic 3D Gaussians and using EWA Splatting~\cite{zwicker2002ewa} with a tiled rasterization pipeline, 3DGS achieves state-of-the-art visual quality at real-time rendering speeds.
However, because each Gaussian has a feature that remains constant throughout its entire volume, capturing high-frequency texture variations requires an excessive number of Gaussians  leading to excessive memory consumption and storage demands.
While recent compression techniques ~\cite{Niedermayr_2024_CVPR, hac++2025} effectively reduce the memory footprint of the Gaussians, they do not address the sheer number of primitives required, which remains a fundamental limitation of per-primitive features.


To improve geometric reconstruction and alleviate volume-like artifacts, 2D Gaussian Splatting ~\cite{huang20242d} constrains the 3D Gaussians into oriented 2D disks(surfels). This improved geometry capture comes at the cost of rendering quality and overfitting capacity and ultimately still struggles to capture high-frequency textures without a large primitive count. 
Other works have proposed generalized, flexible kernels, such as Exponential Gaussians\cite{hamdi2024ges} and Beta-distribution kernels~\cite{liu2025deformable}. 
These flexible kernels enable primitives to better fit high-frequency effects in the scene's structural geometry and sharp boundaries, thus improving rendering quality. However, because they model variations in geometry rather than in texture, they do not fundamentally reduce the number of primitives required to capture textures.

\textbf{\textit{Texture Learning in Radiance Fields}}
Disentangling geometry from appearance is crucial to overcoming these limitations. In the NeRF domain, methods such as NeuTex~\cite{xiang2021neutex} demonstrated the benefits of explicit texture learning by mapping neural features to surface meshes to recover high-frequency details.

In the context of 3D Gaussians and other primitives, the standard formulation relies on storing high-degree Spherical Harmonics (SH) coefficients per primitive, limiting the resolution of surface details to the splat density. To break this dependency, there has been a resurgence of explicit texture mapping for surfel-based representations. By constraining Gaussians to align with the surface(e.g. via 2DGS) and assigning UV coordinates, several methods explicitly map textures onto the primitives ~\cite{weiss2024gaussianbillboardsexpressive2d, svitov2025billboardsplattingbbsplatlearnable, song2024hdgs, xu2024SuperGaussians, rong2024gstex, chao2025texturedgaussians}. Such approaches successfully decouples geometric resolution from appearance resolution, allowing the primitives to render much faster while improving visual quality, but can suffer from unstable optimization or memory restrictions from per-primitive textures.

\textbf{\textit{Hybrid Neural Texturing}}
To avoid memory constraints and optimization instability, Neural Shell Texture Splatting(NeST) ~\cite{zhang2025neural} introduced a highly effective hybrid representation. NeST combines 2D Gaussians with a continuous, multi-resolution spatial hash grid. By fully disentangling the geometry into Gaussians and the appearance on the spatial hash grid, NeST efficiently captures spatial textures with stable optimization and a sparser reconstruction than primitive-based methods. NeST, however, suffers from limitations in its rendering efficiency and loses the tendency of 2DGS to closely model scene geometry due to its per-primitive features. Instead of fully separating features from the geometry, our method strikes a balance between per-primitive features and spatially varying texture-features.

\begin{figure*}[t]
    \centering
    \begin{minipage}[t]{0.65\textwidth}
        \vspace{0pt}
        \includegraphics[width=\textwidth]{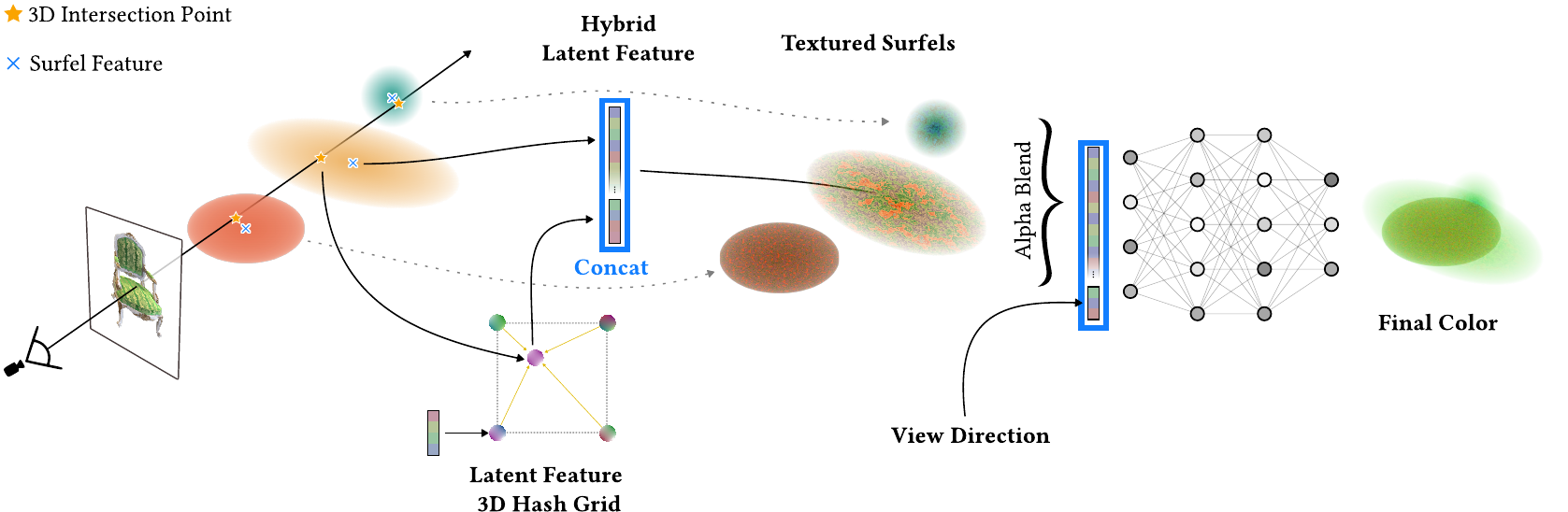}
    \end{minipage}
    \hfill
    {\color{black!20}\vrule width 0.5pt} 
    \hfill
    \begin{minipage}[t]{0.33\textwidth}
        \vspace{0pt}
        \includegraphics[width=\textwidth]{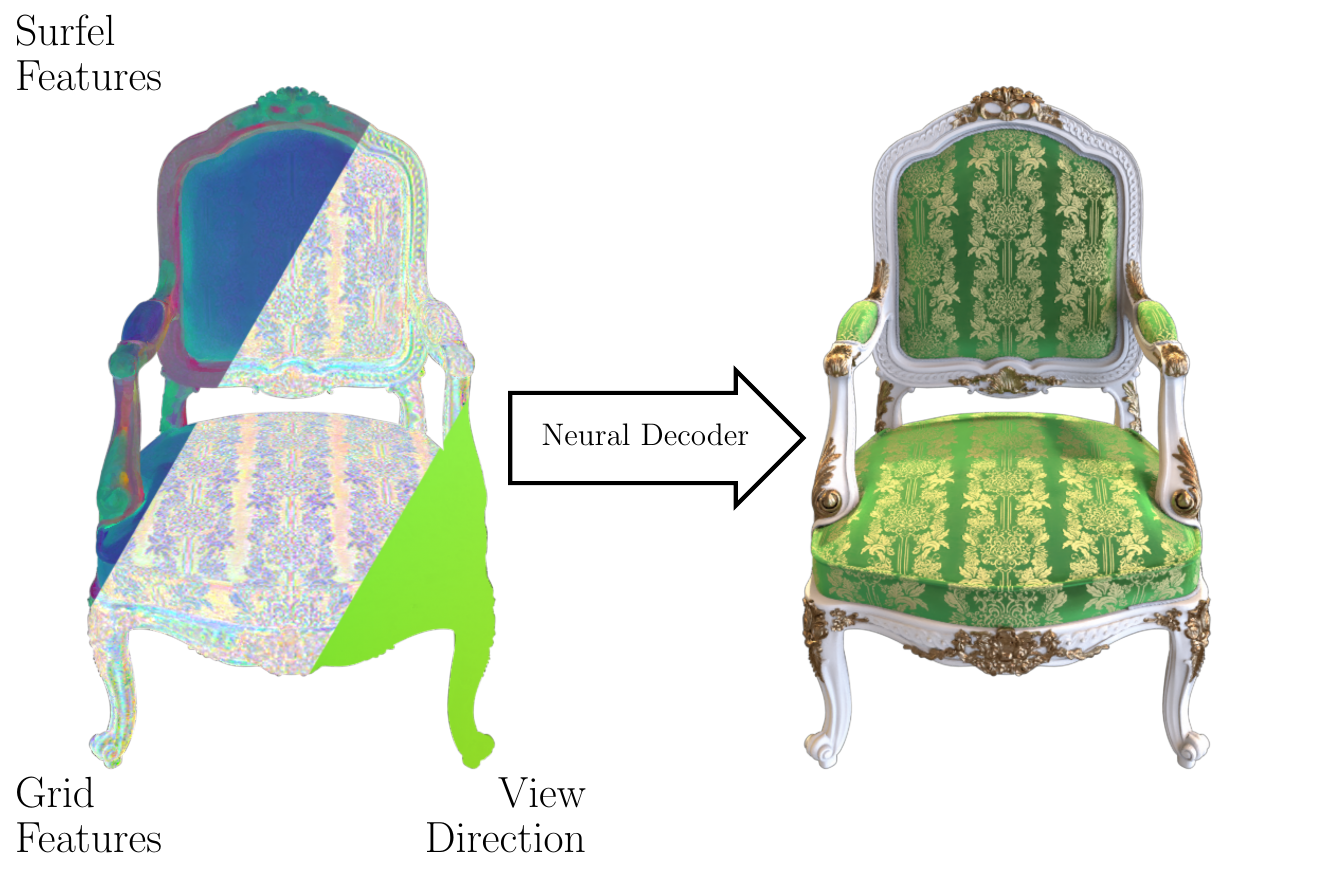}
    \end{minipage}
    \caption{\textbf{Method Overview:} Features from per-surfel representations and a single-resolution hash-grid are blended via volumetric rendering along each view ray. The blended feature, augmented with viewing direction, is processed by an MLP to output color.}
    \label{fig:overview}
\end{figure*}

\section{Method}
Our method augments 2D surfels with a learnable per-surfel latent and modulates this representation with spatial samples from a 3D hash-grid latent. In this way, a strong separation between base geometry - encoded in the per-surfel latent - and residual texture - encoded in the single-resolution hash-grid - is achieved. During rendering, modulated per-fragment latents are blended and decoded by an MLP into color and opacity. Fig. \ref{fig:overview} illustrates the hybrid latent representation and its use in the rendering process.


\subsection{Differentiable Surfel Splatting}

Surfels~\cite{pfister2000surfels}, short for surface elements, are two-dimensional primitives used in computer graphics to represent 3D objects as a dense collection of points rather than a connected polygonal mesh. 
Each surfel typically stores attributes such as position, normal vector, and color, enabling the efficient rendering of complex geometries using point-based techniques.
3DGS~\cite{kerbl_3d_2023} evolves the concept of surfels by replacing flat disks with volumetric ellipsoids, using EWA Splatting to efficiently transform the 3D Gaussians into the 2D image plane for rendering.
2DGS "flattens" the volumetric ellipsoids back into oriented planar disks, effectively returning to the surfel concept but using differentiable, soft-edged Gaussians to ensure the scene remains both geometrically accurate and photorealistic.

Each surfel $i$ is parameterized by a position $\mu_i$, rotation $q_i$, two-dimensional scale $s_i$ and opacity $o_i$.
In differentiable surfel rendering, the color of a pixel $C$ is computed by alpha-blending a depth-sorted sequence of $N$ surfels that overlap a specific screen coordinate.
For any given kernel $G$ (such as a 2D disk, an EWA-filtered ellipsoid, or a 2D Gaussian), the contribution of the $i$-th surfel is determined by its opacity $\alpha_i$ and its spatial influence $G(x)$, leading to the "over" composition formula:
\begin{equation}
    C = \sum_{i=1}^{N} c_i(x) \sigma_i \prod_{j=1}^{i-1} (1 - \sigma_j)
\end{equation}
where $\sigma_i = \alpha_i G(x)$ represents the effective transparency at the point $x$.
The color $c_i(x)$ can either be a per-surfel constant color value (e.g., as in 3DGS, 2DGS), or depend on the position $x$ for textured surfels.

Because this formulation is fully differentiable with respect to the surfel attributes like position, color, and the shape parameters defining $G$, the entire scene can be optimized via differentiable rendering to match a set of reference images as described by Kerbl etal.~\cite{kerbl_3d_2023}.

\begin{figure}[ht!]
\vspace{-2em}
    \centering
    \begin{subfigure}[t]{0.48\linewidth}
        \begin{minipage}[t][5cm][c]{\linewidth}
            \centering
            \includegraphics[width=\linewidth]{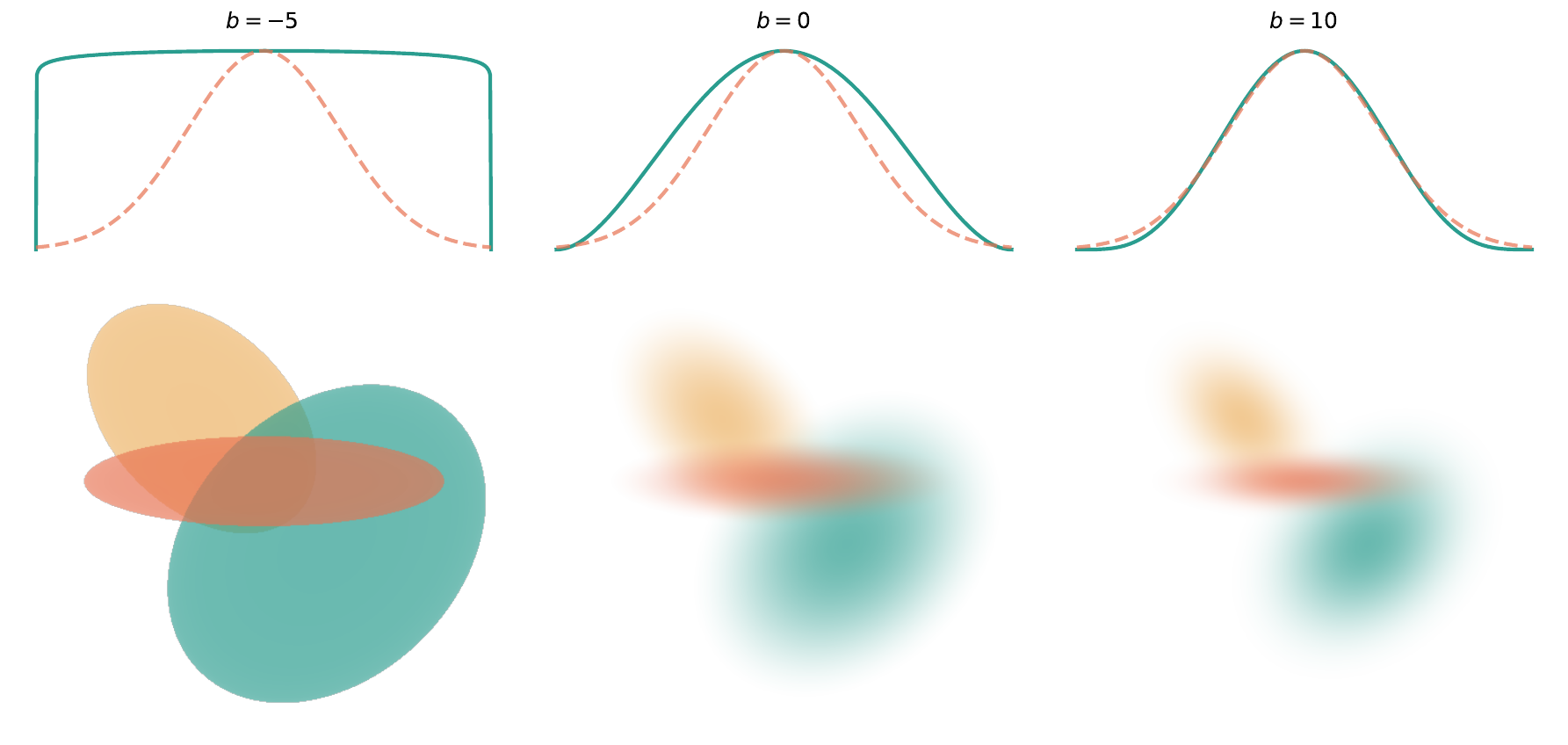}
        \end{minipage}
        \caption{\textbf{Beta kernels}. Negative $b$ values yield flatter peaks with sharper cutoffs for learning sharp geometry, whereas positive $b$ values produce smooth distributions resembling the Gaussian kernel (dashed line).}
        \label{fig:beta-kernels}
    \end{subfigure}
    \hfill
    \begin{subfigure}[t]{0.48\linewidth}
        \begin{minipage}[t][5cm][c]{\linewidth}
            \centering
            \includegraphics[width=\linewidth]{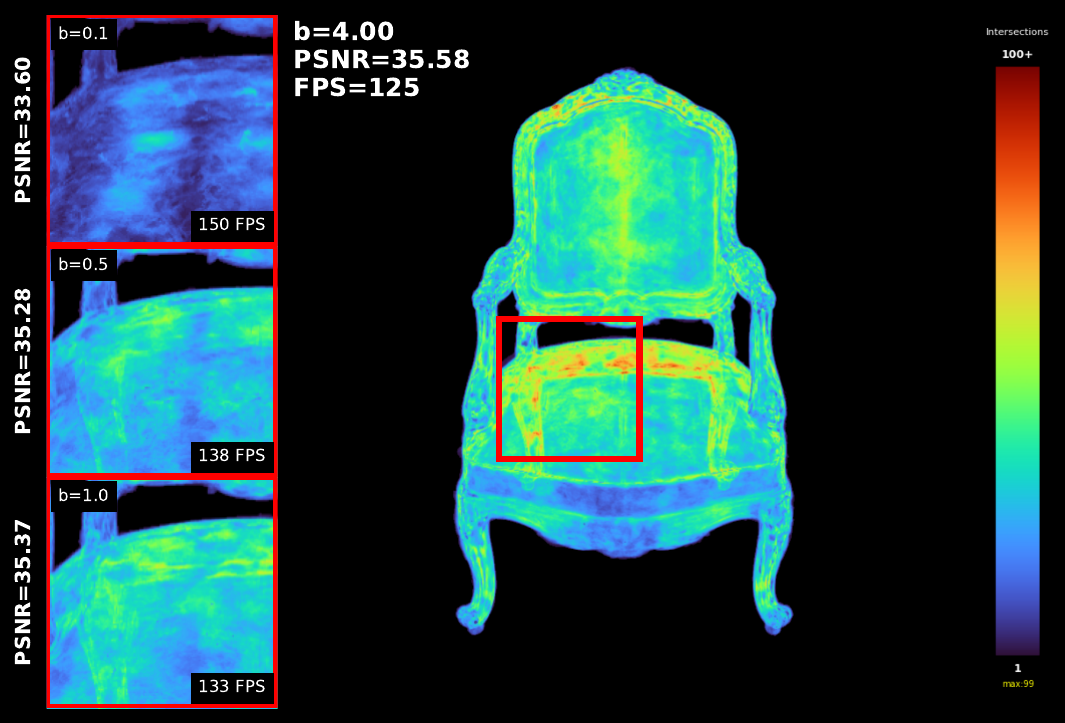}
        \end{minipage}
        \caption{\textbf{Number of rendered surfels contributing to each pixel}. Decreasing $b$ results in decreasing number of surfels, less intersections and increasing frame rates}
        \label{fig:intersection-ablation}
    \end{subfigure}
    \vspace{-1em}
    \caption{\textbf{Beta kernels and their effects.}
    }
    \vspace{-3em}
    \label{fig:combined-analysis}
\end{figure}

\subsection{Deformable beta kernels}
\label{sec:beta-kernels}
Deformable beta kernels \cite{liu2025deformable} replace the Gaussian kernel of 3DGS with a Beta function that has a learnable parameter $b$, allowing it to represent both Gaussian-like and opaque ellipsoidal shape functions. 
We replace the Gaussian kernel used in 2DGS with a beta kernel
\begin{equation}
    \mathcal{B}(x;b) = (1-x)^{\beta{b}},\quad \beta(b)=4\sigma(b),\quad x \in [0,1], b \in \mathbb{R},
    \label{eq:beta-kernel}
\end{equation}
where $\sigma(\cdot)$ is the sigmoid function. $x$ is the normalized radial distance from the center of the kernel.
As shown in Figure~\ref{fig:beta-kernels}, this allows the surfels to represent both 2D Gaussian-like elements (with large $b$ values) while also supporting more disk-like elements with small $b$ values. 
This added flexibility facilitates more efficient surface modeling with fewer primitives, without losing the Gaussians' ability to model non-surface-like structures. As shown in Fig. \ref{fig:intersection-ablation}, flat opaque surfaces can be captured with reduced overdraw, enabling faster rendering.

\subsection{Hybrid Latent Colors}
\label{sec:rendering}

Unlike prior methods relying on a single neural field for representing color, we concatenate a learnable, per-surfel constant feature vector $\mathbf{f}_g \in \mathbb{R}^{D_g}$ 
with a high-frequency latent feature vector $\mathcal{H}(\mathbf{x}_i)$ $\mathbf{f}_h \in \mathbb{R}^{D_h}$, where $\mathbf{x}_i$ is the exact 3D intersection point of the view ray with the surfel plane using the ray-splat intersection formulation from~\cite{Huang2DGS2024} and $\mathcal{H}(\mathbf{x}_i)$ is queried from the neural hash-grid as described by Müller \etal~\cite{muller_instant_2022}. 

\paragraph{Feature Compositing.}
For a given pixel, we iterate through the depth-sorted surfels intersecting the line of sight. For each primitive $p_i$, we first retrieve the learnable base feature $\mathbf{f}_{g,i}$. We then find the per-intersection high-frequency hash feature $\mathcal{H}(\mathbf{x}_i)$ and concatenate it with $\mathbf{f}_{g,i}$. 

Unlike standard splatting, which accumulates view-dependent RGB colors directly, we accumulate the concatenated hybrid feature vectors. The per-pixel feature vector $\mathbf{F}_{pix}$ is computed via alpha blending:

\begin{equation}
    \mathbf{F}_{pix} = \sum_{i \in \mathcal{N}} T_i \alpha_i \cdot \text{concat}\left(\mathbf{f}_{g,i}, \mathcal{H}(\mathbf{x}_i)\right)
\end{equation}

where $T_i$ is the transmittance and $\alpha_i$ is the opacity computed via the beta kernel (Eq.~\ref{eq:beta-kernel}). This late-fusion approach allows the rasterizer to efficiently blend the low-frequency geometric signal ($\mathbf{f}_g$) with the high-frequency texture signal ($\mathbf{f}_h$) in a unified latent space.

\paragraph{Neural Decoding.}
The composited feature vector $\mathbf{F}_{pix}$ acts as a neural descriptor for the surface visible at that pixel. This vector is passed through a MLP, $\Phi$, to recover the final color. To model view-dependent effects such as specularity, the normalized view direction $\mathbf{d}$ is encoded using SH coefficients and concatenated with the feature vector:

\begin{equation}
    \mathbf{C}_{final} = \Phi_{\theta}\left( \mathbf{F}_{pix}, \text{SH}(\mathbf{d}) \right).
\end{equation}

By training the MLP on these concatenated features, the network effectively learns to treat the hash features as a high-frequency residual to the structural ``base paint'' provided by the per-surfel features. This inductive bias enables our method to utilize large surfels as textured billboards, maintaining high visual fidelity even when the primitive count is significantly lowered.




\section{Optimization}
\label{sec:optimization}

Optimizing hybrid representations faces the problem of competing gradients between geometric primitives and the neural texture field.
Gradient-based densification techniques \cite{kerbl_3d_2023} tend to result in geometric dilation, where primitives expand to maximize the neural field's sampling area rather than adhering to the physical surface.
To resolve this, we adopt a stochastic optimization strategy inspired by recent advances in sampling-based rendering.

\subsection{Stochastic Geometry Optimization}

To avoid local minima and better explore the parameter space than pure gradient descent, we follow Kheradmand et al.~\cite{kheradmand20243d}(MCMC) and use Stochastic Gradient Langevin Dynamics to update geometric parameters.
Instead of direct energy minimization, the optimization of 2D Gaussian surfel parameters is framed as sampling from a probability distribution where the density is proportional to rendering quality.

Instead of the heuristic cloning and splitting strategies used in standard 3DGS~\cite{kerbl_3d_2023}, the set of primitives are viewed as particles in a Markov Chain. The update rule for a primitive's position $\mu$ at step $t$ is given by
\begin{equation}
    \mu_{t+1} = \mu_t - \eta \nabla_\mu \mathcal{L}_{total} + \sqrt{2\eta} \cdot \epsilon_t,
    \label{eq:sgld_update}
\end{equation}
where $\eta$ is the learning rate and $\epsilon_t \sim \mathcal{N}(0, \mathbf{\Sigma}_p)$ is noise injected to encourage exploration. Unlike the original 3D formulation, we constrain the noise injection $\epsilon_t$ primarily to the tangent plane of the 2D surfel to encourage surface exploration while minimizing off-surface floating artifacts. This stochasticity allows primitives to escape local minima and ``relocate'' to regions of high reconstruction error without complex densification heuristics.

\subsection{Sparsity via Binary Entropy Regularization}

While the MCMC framework naturally manages primitive density, standard opacity regularization (L1) often leads to a ``fog'' of semi-transparent Gaussians. In our hybrid architecture, this is detrimental for two reasons: (1) it increases the number of expensive neural field queries per ray, and (2) it prevents the formation of the hard surfaces required for efficient adaptive culling.

To enforce the formation of a hard shell, we introduce a Binary Cross-Entropy (BCE) regularization term on the opacity $\sigma_i$. Unlike L1 regularization which encourages opacities to shrink towards zero, BCE encourages opacities to commit to being either fully opaque ($\sigma_i=1$) or fully transparent ($\sigma_i=0$):
\begin{equation}
    \mathcal{L}_{BCE} = \lambda_{bce} \sum_{i} - \left( \sigma_i \log(\sigma_i) + (1-\sigma_i) \log(1-\sigma_i) \right)
    \label{eq:bce_loss}
\end{equation}
By penalizing intermediate values ($0 < \sigma_i < 1$), we force the optimization to make discrete decisions about the placement of primitives. 
Primitives that cannot justify being fully opaque are driven to zero and pruned.
This regularization works in tandem with the beta kernels (Sec.~\ref{sec:beta-kernels}) to produce a scene representation that is both geometrically sharp and highly sparse.

We activate BCE regularization in the later stages of training, after the end of the MCMC relocation phase, to avoid pruning too early and to avoid conflicts with the opacity regularizer in the early stages of optimization.

\subsection{Total Loss}
The final loss function combines the photometric reconstruction loss with the geometric regularizers:
\begin{align}
    \mathcal{L}_{total} &= \mathcal{L}_{RGB} + \lambda_{dist}\mathcal{L}_{dist} \\ &+\lambda_{normal}\mathcal{L}_{normal} + \lambda_{opacity}\mathcal{L}_{opacity}  \\ &+\lambda_{bce}\mathcal{L}_{BCE}
\end{align}
where $\mathcal{L}_{RGB}$ is the standard L1+SSIM rendering loss, and $\mathcal{L}_{dist}$ and $\mathcal{L}_{normal}$ are the distortion and normal consistency losses adapted from 2DGS~\cite{huang20242d}.  $\mathcal{L}_{opacity}$ is the MCMC opacity loss. $\mathcal{L}_{BCE}$ is the sparsity inducing BCE loss.

\section{Experiments}
\label{sec:experiments}

We evaluate our hybrid latent representation on standard benchmarks to demonstrate high rendering quality, geometric reconstruction, and significant sparsity in the number of primitives used. 

\subsection{Implementation Details} 
Our framework is implemented in PyTorch with custom CUDA rasterization.
We build upon the NeST Splatting codebase. 
NeST uses a multi-resolution hash-grid with $L=6$ levels and 4 features per level.
We use a single-level hash-grid with a table size of $2^{19}$ for small scenes (synthetic, DTU) and $2^{21}$ for large scenes (MipNeRF360).
Our hybrid concatenated features consist of 20 hash-grid features and 4 per-surfel features, being on par with the 24 features NeST uses. We use the same MLP architecture as NeST, with two hidden layers of width 256.
We set the initial beta kernel shape parameter to 10 (a Gaussian-like curve) and do the same for any cloned or relocated primitives. We optimize for 30,000 iterations, with a 10,000-iteration 2DGS warm-up phase for initialization. We set our opacity regularization weight $\lambda_{opacity}$ to 0.01 and our BCE regularization weight $\lambda_{BCE}$ to 0.01. The rest of our hyperparameters follow NeST-Splatting. All experiments were conducted on a single NVIDIA RTX 5090 GPU.

\subsection{Datasets and Comparisons}
The evaluation includes the {NeRF Synthetic}~\cite{mildenhall_nerf_2021} dataset, {Mip-NeRF 360}~\cite{barron_mip-nerf_2022} dataset,  and {DTU}~\cite{jensen2014large} dataset. We measure our visual quality using the standard metrics: PSNR, SSIM, and LPIPS used in novel-view synthesis. We compare our method against baseline splatting approaches 3DGS \cite{kerbl_3d_2023} and 2DGS \cite{huang20242d}, adaptive kernel with high/low frequency separation Beta Splatting \cite{liu2025deformable}, per-primitive texturing approach SuperGS \cite{xu2024SuperGaussians}, and the spatial texturing baseline NeST Splatting \cite{zhang2025neural}. 

\subsection{NVS \& Efficiency}
\newcommand{\first}[1]{\cellcolor{red!25}\textbf{#1}}
\newcommand{\second}[1]{\cellcolor{orange!25}\textbf{#1}}
\newcommand{\third}[1]{\cellcolor{yellow!35}\textbf{#1}}

\begin{table*}[ht!]
    \centering
    \caption{\textbf{Quantitative Results.} Comparisons on the NeRF Synthetic, Mip-NeRF 360, and DTU datasets. We include a version of our model with Gaussian kernels and reduced sparsity but higher visual quality. Both kernels demonstrate excellent perceptual quality at reduced primitive counts.}
    \label{tab:nvs_results}
    \resizebox{\textwidth}{!}{
    \begin{tabular}{l|cccc|cccc|cccc}
        \toprule
        & \multicolumn{4}{c|}{\textbf{NeRF Synthetic}} & \multicolumn{4}{c|}{\textbf{Mip-NeRF 360}} & \multicolumn{4}{c}{\textbf{DTU}} \\
        \textbf{Method} & \textbf{PSNR} $\uparrow$ & \textbf{SSIM} $\uparrow$ & \textbf{LPIPS} $\downarrow$ & \textbf{Points} $\downarrow$ & \textbf{PSNR} $\uparrow$ & \textbf{SSIM} $\uparrow$ & \textbf{LPIPS} $\downarrow$ & \textbf{Points} $\downarrow$ & \textbf{PSNR} $\uparrow$ & \textbf{SSIM} $\uparrow$ & \textbf{LPIPS} $\downarrow$ & \textbf{Points} $\downarrow$ \\
        \midrule
        3DGS & 33.34 & \third{0.969} & \first{0.030} & 288k & \third{27.21} & \second{0.815} & 0.214 & 2.7M & 33.58 & \third{0.965} & \third{0.045} & 323k \\
        2DGS & 33.15 & 0.968 & 0.034 & 102k & 27.04 & \third{0.805} & 0.252 & 2.0M & 33.53 & \third{0.965} & 0.050 & 129k \\
        SuperGS & \second{33.71} & \second{0.970} & \second{0.031} & 207k & 26.55 & 0.767 & 0.293 & \third{0.5M} & \second{34.03} & \first{0.967} & 0.055 & 493k    \\
        NeST-Splatting & 33.50 & 0.967 & \third{0.032} & 73k & 26.68 & 0.795 & \third{0.212} & 1.0M & 33.67 & 0.964 & \first{0.042} & \third{80k}    \\
        Beta-Splatting & \first{33.82} & \first{0.971} & \second{0.031} & 100k & \first{28.10} & \first{0.829} & \first{0.192} & 3.1M & \first{34.05} & \second{0.966} & 0.049 & 100k      \\
        Beta-Splatting(small) & \third{33.70} & \third{0.969} & 0.035 & \third{50k} & \second{27.28} & 0.795 & 0.259 & \second{0.4M} & \third{33.93} & \third{0.965} & 0.058 & \first{50k}      \\
        \midrule
        {Ours (Gaussian)} & {33.53} & \third{0.969} & \second{0.031} & \second{36k} & {27.20} & {0.801} & \second{0.199} & {0.7M} & {33.73} & \third{0.965} & \first{0.042} & 95k \\
        {Ours (Beta)} & {33.20} & {0.954} & {0.035} & \first{19k} & {26.85} & {0.794} & {0.218} & \first{0.2M} & {33.63} & {0.963} & \third{0.045} & \first{50k} \\
        \bottomrule
    \end{tabular}
    }
\end{table*}

We report PSNR, SSIM, LPIPS, and the number of primitives(Points) in Table~\ref{tab:nvs_results}. Our method with Gaussian kernels achieves high rendering quality, while integrating beta kernels yields significantly sparser reconstructions (at higher framerates - Table~\ref{tab:bicycle_efficiency}).
We attribute the LPIPS improvement to the hash-grid's effective capture of scene textures. This complements our per-primitive features, which capture larger effects and thus improve PSNR. 

We note that our primary objective in using beta kernels is to reduce redundant hash queries, whereas Beta Splatting relies on a large number of semi-transparent primitives to achieve high visual fidelity. Upon fixing the number of 3D beta kernels to 0.4M, we achieve superior LPIPS with even fewer primitives. Beta Splatting reports higher PSNR due to its usage of spherical harmonics, and it also benefits from using 3D kernels compared to our 2D surfels, as flat primitives are more optimized for geometry reconstruction than visual quality. This effect is evident in the superior visual quality of 3DGS compared to 2DGS.

\subsection{Spectral Disentanglement Analysis}

\begin{figure}[ht!]
    \centering
    \includegraphics[width=\linewidth]{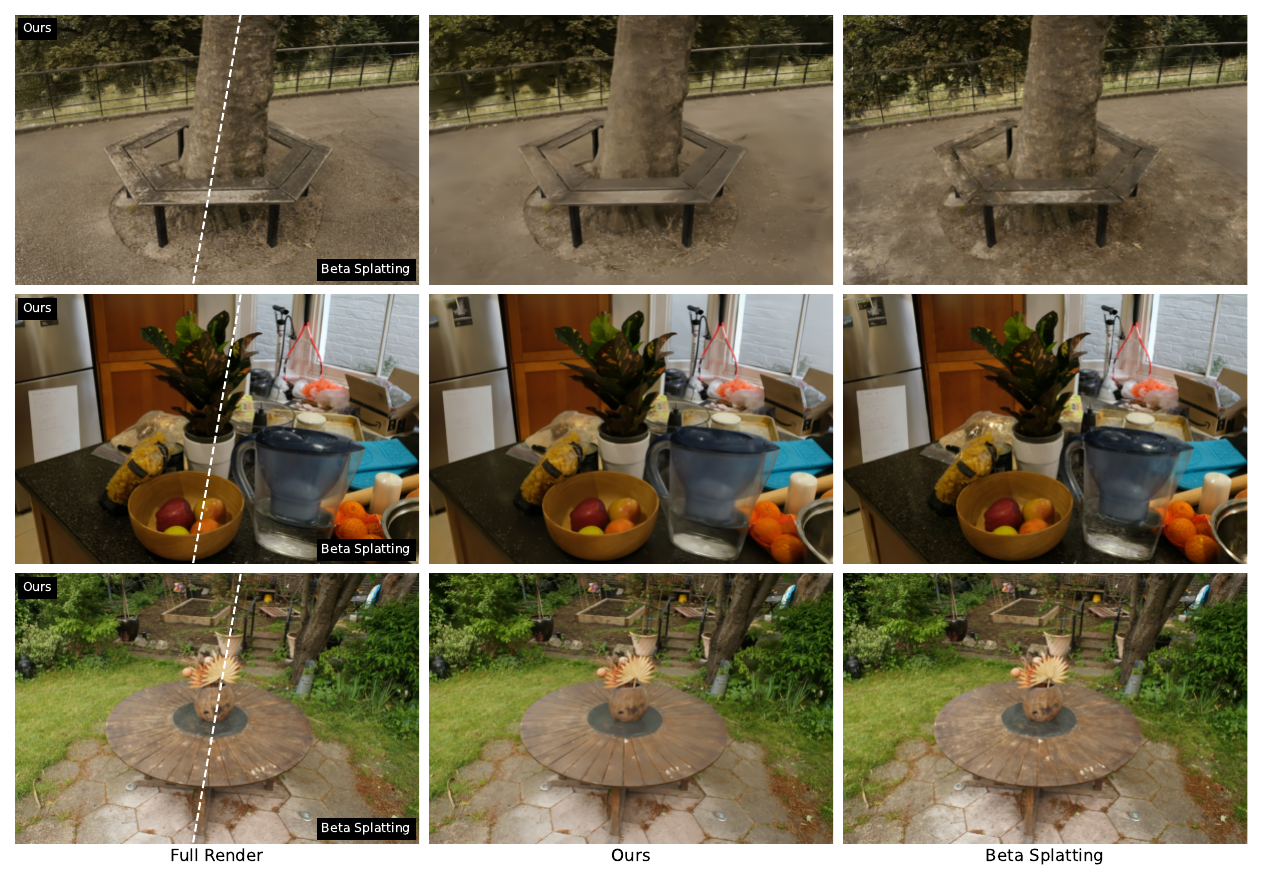}
    \caption{\textbf{Qualitative comparison of Beta Splatting~\cite{liu2025deformable} and Ours}. Left: Full rendering. Middle: Our without the hash-grid to visualize the low-frequency components encoded in the surfel latents. Right: Beta Splatting using the 70\% of splats with the lowest $b$ values, as proposed by the authors for low-frequency decomposition.}
    \label{fig:disentanglement}
\end{figure}

A key contribution of our method is the semantic separation of low- and high-frequency scene components.
To demonstrate this, we visualize the components of our trained model in Figure~\ref{fig:disentanglement}. We render the scene while isolating our per-primitive features by masking out the hash-grid contribution. The result is a smooth, geometry-consistent reconstruction that captures large lighting effects and base colors (inadvertently low-frequency), confirming that our primitives adhere to the scene structure.

In contrast, Beta Splatting\cite{liu2025deformable} proposes to mask out primitives with high beta values to extract the structural components of the scenes.

\begin{figure}[ht!]
\vspace{0em}
    \centering
    \includegraphics[width=\linewidth]{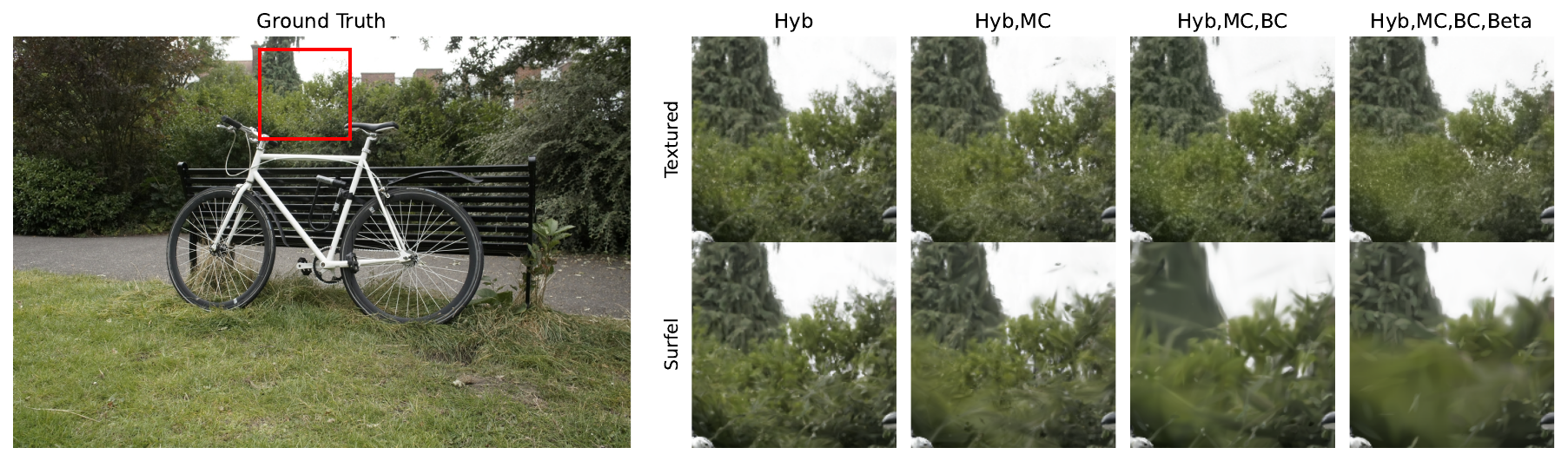}
    \caption{\textbf{Per-primitive information with increasing sparsity}. Left: GT Image. Top: Renders with hybrid features. Bottom: Renders with the hash-grid features disabled. With each additional step in Fig. \ref{tab:bicycle_efficiency}, surfels turn more into textured billboards.}
    \label{fig:billboards}
    \vspace{-1em}
\end{figure}

We see that this often results in broken geometry or empty voids, since their separation is simply a byproduct of lower beta values that favor low-frequency effects, but there is no explicit separation of the two. Our hybrid latents share the per-primitive feature at all intersection points over a surfel, creating an inductive bias towards covering lower-frequency effects. When combined with sparsity regularization, this makes the primitives as large as possible and ensures they cover the scene geometry without holes. Furthermore, since we query the hash-grid at each 3D intersection, it naturally captures the high-frequency effects in the scene, effectively acting as textured billboards (see Fig~\ref{fig:billboards}).

    
    
    
    

\begin{figure*}
    \includegraphics[width=\textwidth]{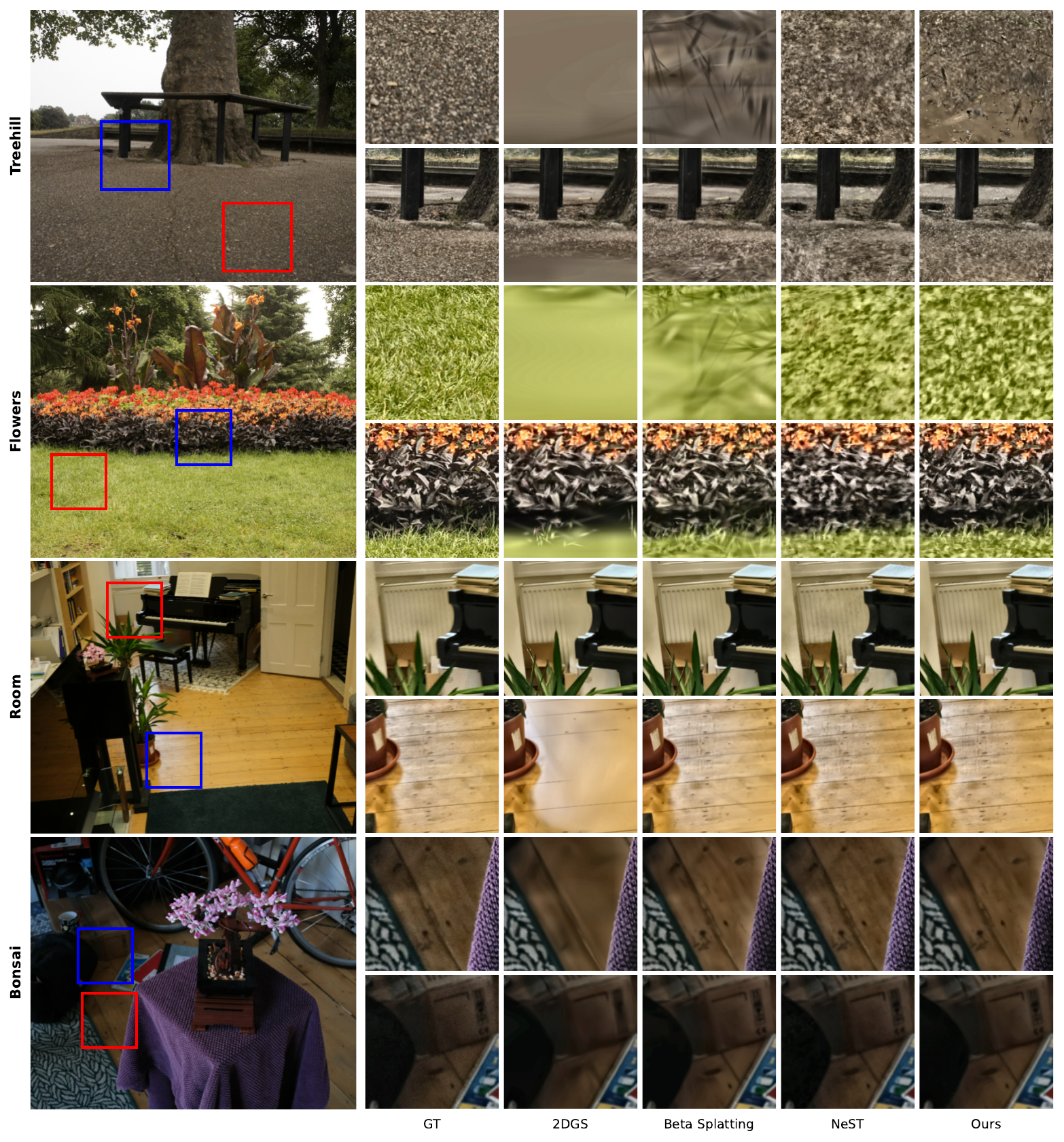}
    \caption{\textbf{Qualitative results} for 2DGS~\cite{Huang2DGS2024}, Beta Splatting~\cite{liu2025deformable}, NeST~\cite{zhang2025neural} and Ours from the MipNeRF Dataset~\cite{barron_mip-nerf_2022}. The images are contrast-enhanced.}
    \label{fig:mipnerf-cutouts}
\end{figure*}

\subsection{Geometric Reconstruction}

We quantify geometric accuracy on the DTU dataset using Chamfer Distance (CD). As shown in Table~\ref{tab:geometry_dtu}, our method achieves lower error than NeST Splatting. By forcing the primitives to carry the low-frequency signal, we stabilize the optimization and obtain better depth and geometry.

\begin{table}[ht!]
    \centering
    \begin{minipage}[t]{0.35\linewidth}
        \centering
        \caption{\textbf{Geometric reconstruction accuracy on DTU} measured by Chamfer Distance (mm). Lower is better.}
        \label{tab:geometry_dtu}
        \resizebox{\linewidth}{!}{%
            \begin{tabular}{lc}
                \toprule
                \textbf{Method} & \textbf{Chamfer Dist. (mm)} $\downarrow$ \\
                \midrule
                3DGS & 1.96 \\
                2DGS & 0.80 \\
                NeST-Splatting & 0.89 \\
                {Ours} & {0.85} \\
                \bottomrule
            \end{tabular}%
        }
    \end{minipage}
    \hfill
    \begin{minipage}[t]{0.61\linewidth}
        \centering
        \caption{\textbf{Reconstruction sparsity on the Bicycle Scene of Mip 360.} We demonstrate increasing sparsity with each additional component.}
        \label{tab:bicycle_efficiency}
        \resizebox{\linewidth}{!}{%
            \begin{tabular}{l|cccc}
                \toprule
                \textbf{Method} & \textbf{PSNR}  $\uparrow$ & \textbf{LPIPS} $\downarrow$ & \textbf{Points} $\downarrow$ & \textbf{FPS} $\uparrow$ \\
                \midrule
                NeST & 24.49 & 0.236 & 2M & 23\\
                Hyb & 24.72 & 0.217 & 3M & 20 \\
                Hyb,MC & 24.77 & 0.234 & 0.76M  & 25 \\
                Hyb,MC,BC & 24.37 & 0.252 & 0.14M & 59 \\
                Hyb,MC,BC,Beta & 23.41 & 0.292 & 0.08M & 80 \\
                \bottomrule
            \end{tabular}%
        }
    \end{minipage}
\end{table}

\subsection{Ablation: Efficiency Analysis}
We analyze the efficiency impact of each step of our method for the Bicycle scene in the Mip-NeRF 360 dataset in Table~\ref{tab:bicycle_efficiency}. When we introduce hybrid Gaussian features into the NeST framework, we see an improvement in visual quality at the cost of sparsity. Compared to NeST, where the per-Gaussian positional gradient is influenced by its multi-resolution hash-grid, our hybrid latents provide more stable gradients due to fewer hash collisions from a single hash layer. The number of primitives increases due to the greater capacity to overfit the scene in per-surfel features, whereas a full hash-grid method is limited by the maximum size of the hash table.

We then introduce MCMC optimization with a maximum budget of 1 million surfels. We observe that the PSNR improves and the number of primitives reduces by a factor of 3. Despite the sparser reconstruction, we see a negligible improvement in framerate because MCMC favors utilizing foggy primitives to maximize visual fidelity. This leads to more intersections per pixel, more hash-grid queries, and reduced rendering speed.

Upon adding BCE regularization, the number of primitives declines to a tenth of the original NeST reconstruction. As we prune a large number of Gaussians from the MCMC optimization, we lose some visual fidelity but see a large increase in the framerate. This is partly due to the reduced number of primitives but mainly because of the reduced number of intersections per pixel.

Finally, when we change our surfels to beta kernels, we obtain a dramatic improvement in efficiency at the cost of rendering quality. As our method is geared towards sparsity, our beta kernels tend to favor flatter reconstructions. High opacities from our BCE and flat beta kernels lead to significantly fewer intersections per pixel, giving us a further rendering speed boost and enabling our method to perform real-time rendering. We attribute the lower reconstruction quality to two factors:
Due to the bounded support of beta kernels and the flat opacity falloff of lower beta values, our kernels receive weaker positional gradients, reducing their capacity to overfit a scene without the exploration component of the MCMC optimization. A sparse number of per-primitive latent features, coupled with a single-level hash grid, results in lower representational capacity.

\section{Limitations}

Despite achieving high reconstruction quality with far fewer surfels, the large view-space MLP with 256-dim hidden layers ends up being a performance bottleneck. This makes it difficult to compete with the simple spherical-harmonic queries of Gaussian splatting methods.

Furthermore, the implicit nature of the hash-grids and the view-space decoder limits applications that require merging different scene models, such as scene editing operations. While explicit primitives can be easily moved or combined, merging hash-grids optimized for separate scenes remains an open challenge.


\section{Discussion and Conclusions}
\label{Discussion}
We demonstrate, through extensive experiments on real-world and synthetic datasets, that our proposed hybrid latent representation achieves a significant improvement in scene reconstruction, effectively bridging the gap between high visual fidelity and sparse reconstruction. 
By combining a sparse hash-grid with differentiable surfel primitives, the model captures fine-grained surface details that are often missed by traditional point-based methods. 
Furthermore, due to its sparsity, our approach proves remarkably efficient and compact; by leveraging latent features to encode complex appearance data, the system requires far fewer primitives and exhibits lower overdraw without sacrificing quality.

In the future, we see substantial potential to further optimize rendering speeds. One promising direction is to shift the decoder to 3D and bake the resulting per-intersection RGB values directly into static, textured surfels. This shift would move the computational burden from real-time latent decoding to standard rasterization pipelines, potentially enabling photorealistic performance on mobile or resource-constrained hardware. 

Another research direction would be to further leverage the geometric accuracy of our approach to reconstruct explicit geometric scene representations, for instance, by reconstructing watertight textured meshes from the surfel representation. Given its properties, our hybrid representation framework could serve as a powerful texturing method for mesh reconstruction techniques such as MeshInTheLoop \cite{milo2025} or TriangleSplatting \cite{Held2025Triangle}.




%
%
\bibliographystyle{splncs04}
\bibliography{main}
\newpage

\section{Supplementary Materials}

\subsection{Sparsification methods}
There have been recent works on sparsifying Gaussian reconstructions, such as Mini-Splatting\cite{fang2024mini} and GaussianSpa\cite{zhang2025gaussianspa}, providing significantly sparser reconstructions at higher quality than the vanilla 3DGS optimization. 
For instance, GaussianSpa is an optimization-based simplification framework, where simplification is formulated as a constrained optimization problem. While GaussianSpa alternately solves two independent sub-problems and then uses classical 3DGS for novel view synthesis, our approach addresses explicitly the disentanglement between geometry and appearance. 

Since in GaussianSpa the feature representation is still per-Gaussian spherical harmonics, our experiments show that below a certain number of Gaussians, such features struggle to capture textures regardless of the optimization method. 

\begin{figure}[ht!]
    \centering
    \includegraphics[width=\linewidth]{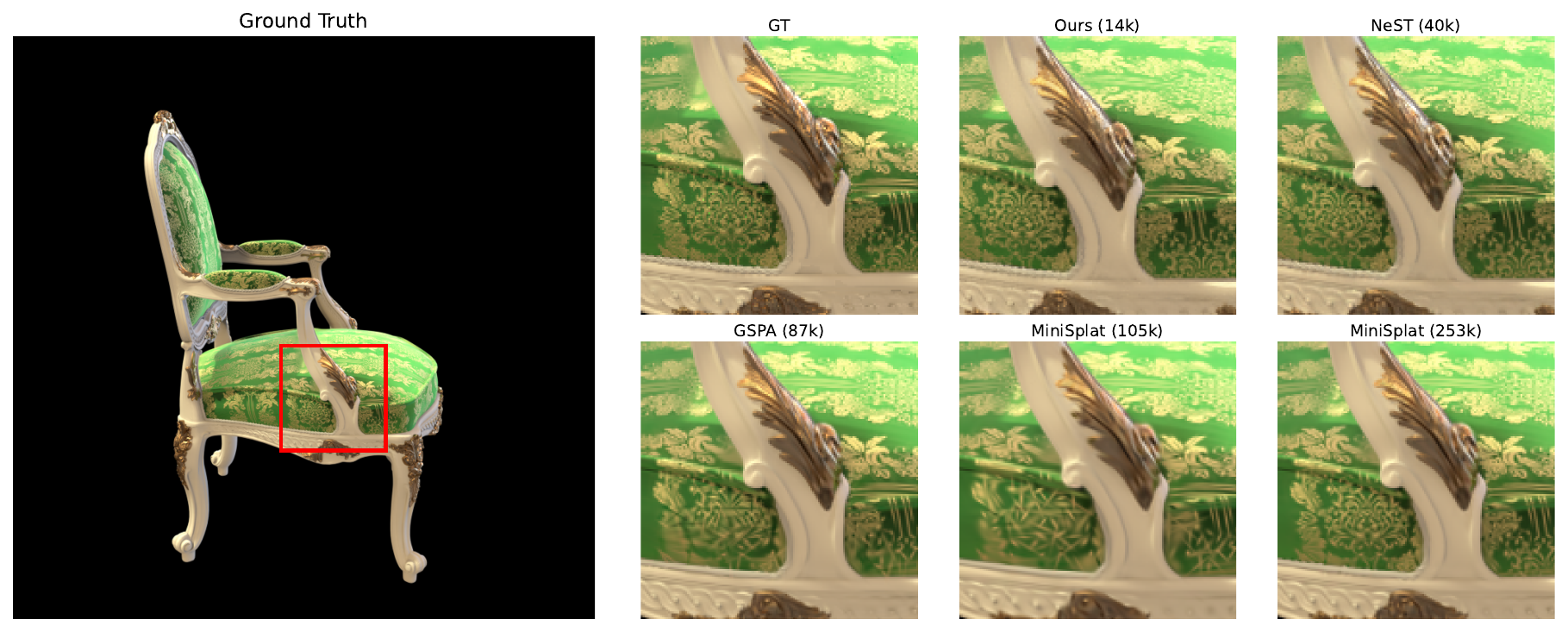}
    \caption{\textbf{Per-primitive features vs texturing methods}. Our method and NeST Splatting manage to represent the textures from the chair scene. GaussianSpa and Mini-Splat both fail below a certain primitive count due to the limitations of per-primitive spherical harmonics.}
    \label{fig:texture-sparsity}
\end{figure}

As we can see in Figure \ref{fig:texture-sparsity}, both GaussianSpa and Mini-Splatting are unable to represent the chair texture at a reduced primitive count. Mini-Splatting still manages to represent it at a higher primitive count, something that the baseline Gaussian Splatting optimization is incapable to do. NeST Splatting and our hybrid representation capture the textures perfectly even at a low primitive count.

We also note that our hybrid latent method can be combined with either of these sparsifying optimizations. We select MCMC in the paper because Beta-Splatting - the method closest to ours regarding the separation of low and high frequency components into the scene geometry - utilizes MCMC for its optimization.

\newpage
\subsection{Hash-grid Levels}
In Table \ref{tab:hash-grid_levels}, we compare the effect of replacing the coarser levels of the multiresolution hash-grid in NeST Splatting with per-Gaussian features. The total feature size is set to 24D. We use 6 hash levels, which is equivalent to the baseline NeST, and as we replace coarser levels, the hybrid feature dimensionality remains at 24D. 

Using fewer hash-levels helps us avoid the overhead of querying a multiresolution hash-grid. Since the per-Gaussian feature is loaded into shared memory, this speeds up training and inference in addition to the speed benefit from the increased sparsity of our method. We even observe improvements in the rendering quality with a hybrid representation.

\begin{figure}[ht]
    \centering
    \includegraphics[width=\linewidth]{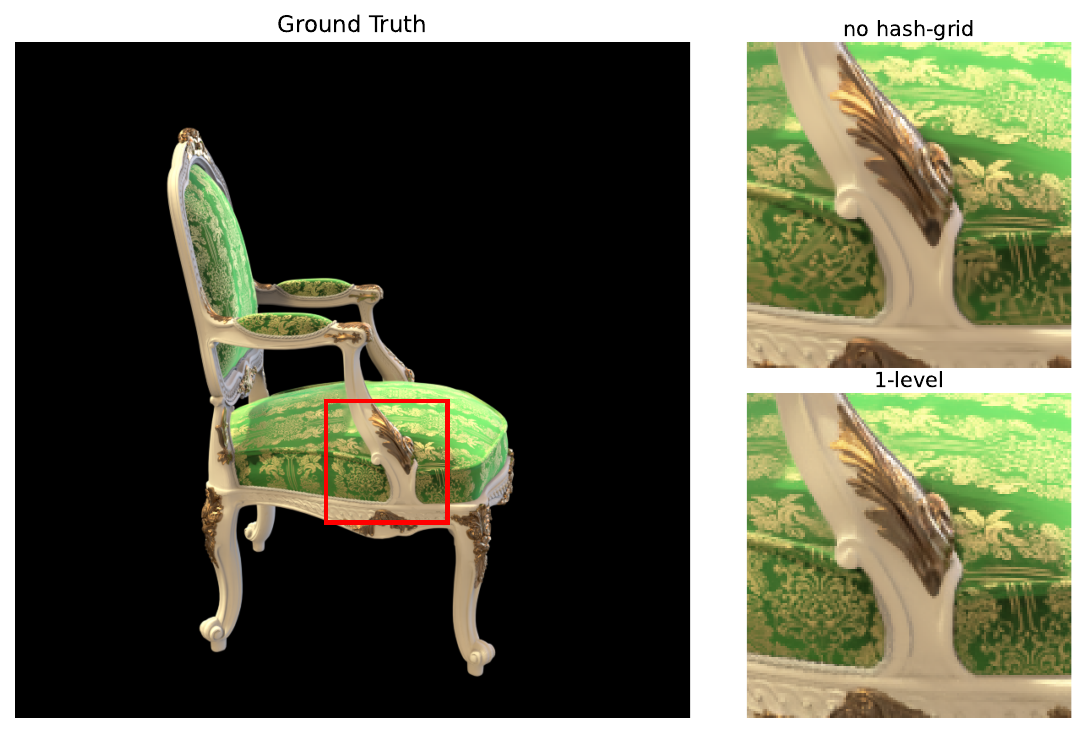}
    \caption{\textbf{Using a single hash-grid layer vs pure per-primitive features}. Using only per-primitive features causes similar artifacts to using per-Gaussian spherical harmonics. A hybrid feature with even a single hash-grid layer captures all texture variation.}
    \label{fig:hash-levels}
\end{figure}

Replacing all hash-grid levels with per-Gaussian features leads to the same limitations in texture representation as Gaussian Splatting, as demonstrated in Figure \ref{fig:hash-levels} (no hash-grid). Per-Gaussian features struggle to represent textures, regardless of whether we use spherical harmonics or a 2D MLP decoder.

\begin{table}[ht]
\centering
\setlength{\tabcolsep}{12pt}
\begin{tabular}{cccc}
\toprule
Hash Levels & Mip360 & NeRFSyn & DTU \\
\midrule
6(NeST) & 26.52 & 33.37 & 33.62 \\
5 & 26.71 & 33.46 & 33.73 \\
4 & 26.64 & 33.49 & 33.75 \\
3 & 26.62 & 33.52 & 33.78 \\
2 & 26.56 & 33.50 & 33.77 \\
1 & 26.62 & 33.50 & 33.69 \\
0 & 26.23 & 33.42 & 33.60 \\
\bottomrule
\end{tabular}
\caption{PSNR scores across different hash-grid levels.}
\label{tab:hash-grid_levels}
\end{table}

\newpage

\subsection{MCMC primitive limits}
For all Beta Splatting results in the novel-view-synthesis table(Tab. 1 in paper) except the large model on the Mip-NeRF 360 dataset, we report the primitive upper-limit set on the MCMC optimization since their formulation relies on utilizing a lot of low-opacity primitives to overfit a scene. The actual active number of primitives (opacity<0.001) might be slightly lower than the limit set by the MCMC optimization. In our method we observe a significant reduction from the cap due to our BCE optimization deleting low opacity Gaussians.

\subsection{Feature Decomposition}
\begin{figure}[ht!]
    \centering
    \includegraphics[width=\linewidth]{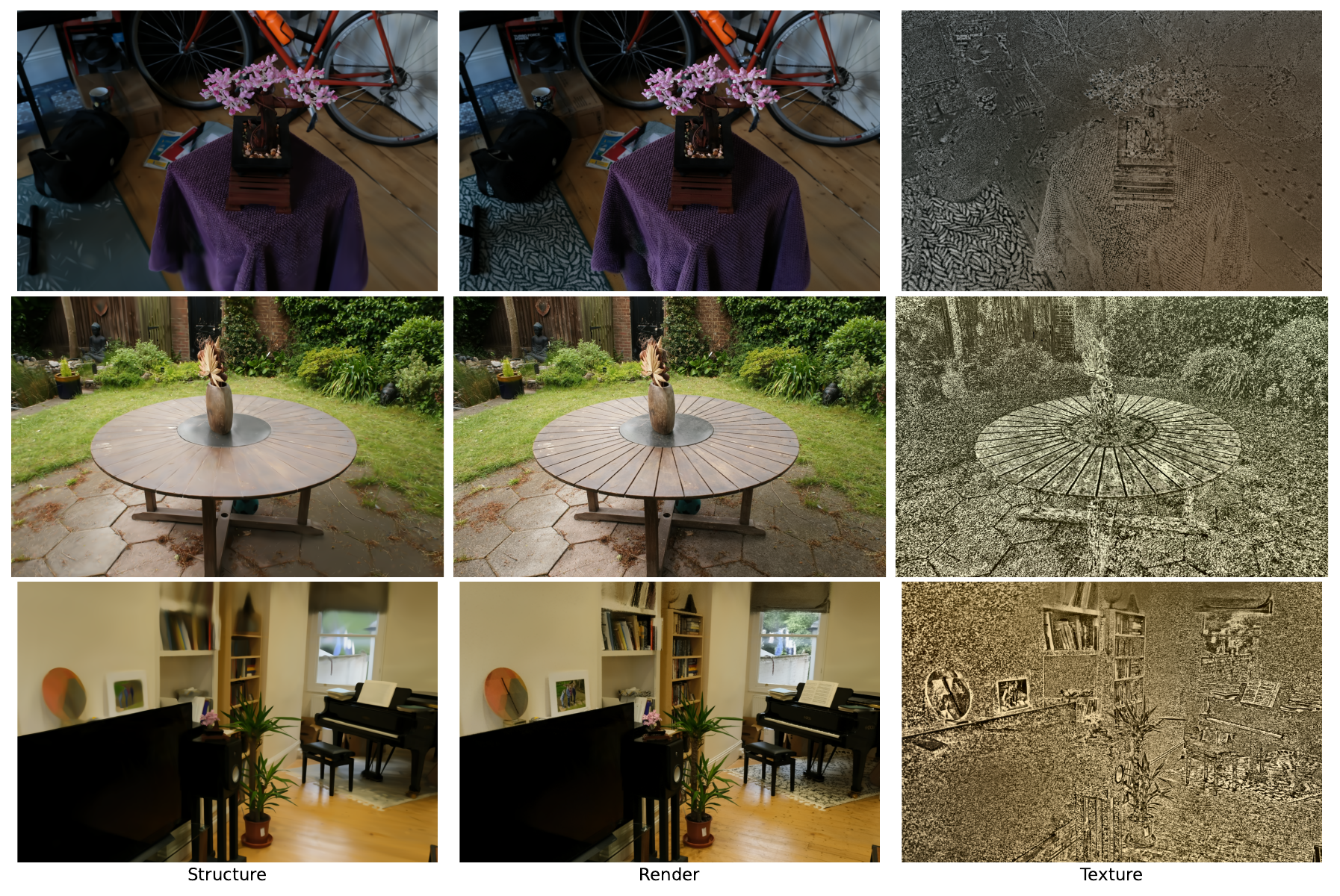}
    \caption{\textbf{The separation of features into low-frequency structural components on the primitives (left) and high-frequency texture components in the hash-grid (right)}.}
    \label{fig:decomposition}
\end{figure}

In Figure \ref{fig:decomposition}, in the first column, we mask out the hash features to obtain the low-frequency structural details of our method. In the middle column we show the combined hybrid feature output. In the third column, we mask out the per-primitive features to visualize what texture details are stored in the hashgrid. We also provide videos of these decompositions.

\subsection{Mip-360}
For the Mip-NeRF 360 dataset, we use the --images flag, which uses the ImageMagick downscaled images as used in the original 3DGS. Using the --resolution flag provides better results with its bicubic downsampling. In Table \ref{tab:mip360} we report our per-scene results for the Mip-NeRF 360 dataset with our sparsest method with the beta kernels, our best looking method with Gaussian kernels but with less sparsity, and our sparsest Gaussian kernels with the same setting as the Beta kernel results.

\begin{table*}[t]
\centering
\caption{Per-scene PSNR, SSIM, LPIPS, and \#Primitives on the Mip-NeRF 360 dataset. }
\label{tab:mip360}
\resizebox{\textwidth}{!}{%
\begin{tabular}{lcccc|c|ccccc|c|c}
\toprule
Method & Bon & Ctr & Kit & Rm & \textbf{Ind.} & Bic & Flw & Gdn & Stp & Trh & \textbf{Out.} & \textbf{Mean} \\
\midrule

\midrule
Beta & 32.13 & 28.60 & 30.45 & 30.44 & 30.41 & 24.45 & 20.15 & 26.94 & 26.27 & \textbf{22.26} & 24.01 & 26.85 \\
Gauss(Large) & \textbf{33.01} & \textbf{29.21} & 30.77 & \textbf{31.53} & \textbf{31.13} & \textbf{24.61} & \textbf{20.52} & 27.04 & \textbf{26.31} & 21.80 & \textbf{24.06} & \textbf{27.20} \\
Gauss(Small) & 32.55 & 28.79 & \textbf{30.85} & 30.69 & 30.72 & 24.48 & 20.49 & \textbf{27.11} & 26.08 & 21.97 & 24.03 & 27.00 \\
\midrule

\midrule
Beta & 0.935 & 0.894 & 0.914 & 0.904 & 0.912 & 0.731 & 0.549 & 0.846 & 0.754 & \textbf{0.619} & 0.700 & 0.794 \\
Gauss(Large) & \textbf{0.943} & \textbf{0.905} & \textbf{0.921} & \textbf{0.913} & \textbf{0.921} & \textbf{0.748} & \textbf{0.574} & \textbf{0.853} & \textbf{0.762} & 0.596 & \textbf{0.707} & \textbf{0.802} \\
Gauss(Small) & 0.938 & 0.898 & 0.919 & 0.907 & 0.916 & 0.736 & 0.571 & 0.850 & 0.750 & 0.617 & 0.705 & 0.79F9 \\
\midrule

\midrule
Beta & 0.192 & 0.208 & 0.134 & 0.209 & 0.186 & 0.234 & 0.329 & 0.130 & 0.234 & 0.292 & 0.244 & 0.218 \\
Gauss(Large) & \textbf{0.182} & \textbf{0.188} & \textbf{0.121} & \textbf{0.197} & \textbf{0.172} & \textbf{0.202} & \textbf{0.298} & \textbf{0.114} & \textbf{0.211} & \textbf{0.281} & \textbf{0.221} & \textbf{0.199} \\
Gauss(Small) & 0.188 & 0.201 & 0.129 & 0.207 & 0.181 & 0.229 & 0.310 & 0.126 & 0.239 & 0.286 & 0.238 & 0.213 \\
\midrule

\midrule
Beta & \textbf{103.20} & \textbf{76.10} & \textbf{123.00} & \textbf{44.30} & \textbf{86.70} & \textbf{328.50} & \textbf{407.80} & \textbf{331.30} & 399.00 & \textbf{294.20} & \textbf{352.20} & \textbf{234.20} \\
Gauss(Large) & 373.00 & 321.90 & 502.70 & 227.30 & 356.20 & 2291.10 & 1688.40 & 1795.00 & 2242.40 & 1948.80 & 1993.10 & 1265.60 \\
Gauss(Small) & 159.80 & 119.40 & 209.50 & 70.40 & 139.80 & 417.30 & 627.80 & 374.90 & \textbf{370.00} & 469.20 & 451.80 & 313.10 \\
\bottomrule
\end{tabular}}
\end{table*}

\subsection{Beta Kernels}
We use beta kernels with a sigmoidal range between 0 and 4 instead of the full exponential range of Beta-Splatting since having a larger value of Beta only makes the kernels sharp with a smaller support. Our method already separates appearance into low and high frequencies so we do not require the sharper Beta kernels for fitting these high-frequency details.

\subsection{2DGS warm-up}
As NeST Splatting, our method needs to start with a 2DGS warm-up phase for the best results, since the initial optimization faces problems with the used screen-space MLP. Methods such as MeRF \cite{reiser2023merf} and SMeRF \cite{duckworth2024smerf} combine a small view-space MLP with a view-independent 3D MLP during training that they bake into a light-weight feature representation during inference. This 3D MLP formulation would be a good future direction to explore for from-scratch optimization and faster inference.

\subsection{Parameter Count}
We do not explore the parameter count of our method as the purpose of this work is sparsity. Since we use a single hash-grid level and our per-Gaussian features are 24D, much smaller than the 48D SH used by Gaussian Splatting, our parameter count should be significantly lower when combined with a post-processing quantization method \cite{Niedermayr_2024_CVPR}. For further storage efficiency one can even quantize the hashgrid using a method such as SHACIRA\cite{girish2023shacira}. The overhead of storing the MLP weights is negligible.

\subsection{Rendering Speed}
Since the purpose of this paper is to demonstrate the effectiveness of a hybrid feature representation, we have not attempted to optimize the rendering pipeline of NeST Splatting. The large width of the MLP hidden layers (256) and the size of the features in our experiments (24D) match the NeST implementation and are bottlenecks in improving rendering speed. As mentioned in the paper, there are various future directions to explore for improving rendering speed. With the success of methods like SMeRF \cite{duckworth2024smerf} and MeRF \cite{reiser2023merf}, modifying the MLP architecture might be one of them.

\end{document}